\documentclass[11pt]{article}

\usepackage{acl}

\usepackage{enumitem}
\usepackage{latexsym}
\usepackage{microtype}
\usepackage{inconsolata}

\usepackage{fontspec}
\usepackage{xeCJK}
\setCJKsansfont{FandolHei-Regular.otf}[BoldFont=FandolHei-Bold.otf]
\newcommand{\zh}[1]{#1}

\usepackage{soul}
\usepackage{url}
\usepackage{graphicx}
\usepackage{amsmath}
\usepackage{amsthm}
\usepackage{booktabs}
\usepackage{algorithm}
\usepackage{algorithmic}
\usepackage{comment}
\usepackage{placeins}
\usepackage{float}
\usepackage{multirow}
\usepackage{subcaption}
\usepackage{longtable}
\usepackage{siunitx}
\usepackage{adjustbox}
\usepackage{tabularx}
\usepackage{array}
\usepackage{colortbl}
\usepackage{xcolor}
\usepackage{xspace}

\title{MATCHA: A Chinese Multi-Disorder Mental Health Dataset with Cross-Lingual Analyses on Social Media}

\author{
\textbf{Bingrui Jin}\textsuperscript{1,*},
\textbf{Baihan Li}\textsuperscript{1,*},
\textbf{Zitong Li}\textsuperscript{1,*},
\textbf{Kunyao Lan}\textsuperscript{1}, \\
\textbf{Shimeng Li}\textsuperscript{1},
\textbf{Kenny Q. Zhu}\textsuperscript{2},
\textbf{Mengyue Wu}\textsuperscript{1}
\\[0.8ex]
\textsuperscript{1}Shanghai Jiao Tong University, Shanghai, China \\
\textsuperscript{2}University of Texas at Arlington, Arlington, TX, USA
\\[0.5ex]
\texttt{\{suzy\_jin, LBH0612, mengyuewu\}@sjtu.edu.cn} \\
\small{
\textsuperscript{*}Equal contribution.\quad
\textbf{Correspondence:}
\href{mailto:mengyuewu@sjtu.edu.cn}{mengyuewu@sjtu.edu.cn}
}
}

\newcommand{\figref}[1]{Figure~\ref{#1}}

\newcolumntype{Y}{>{\raggedright\arraybackslash}X}

\definecolor{myOrange}{RGB}{255,150,50}
\definecolor{myGreen}{RGB}{60,160,120}

\newcommand{\colorcell}[1]{%
  \begingroup
  \def\val{#1}%
  \ifdim \val pt > 0pt
    \ifdim \val pt > 3.0pt \cellcolor{myOrange!100}\val%
    \else\ifdim \val pt > 2.5pt \cellcolor{myOrange!90}\val%
    \else\ifdim \val pt > 2.0pt \cellcolor{myOrange!80}\val%
    \else\ifdim \val pt > 1.5pt \cellcolor{myOrange!70}\val%
    \else\ifdim \val pt > 1.0pt \cellcolor{myOrange!60}\val%
    \else\ifdim \val pt > 0.7pt \cellcolor{myOrange!50}\val%
    \else\ifdim \val pt > 0.5pt \cellcolor{myOrange!40}\val%
    \else\ifdim \val pt > 0.3pt \cellcolor{myOrange!30}\val%
    \else\ifdim \val pt > 0.2pt \cellcolor{myOrange!25}\val%
    \else\ifdim \val pt > 0.1pt \cellcolor{myOrange!20}\val%
    \else\cellcolor{white}\val%
    \fi\fi\fi\fi\fi\fi\fi\fi\fi\fi
  \else\ifdim \val pt < 0pt
    \ifdim \val pt < -3.0pt \cellcolor{myGreen!100}\val%
    \else\ifdim \val pt < -2.5pt \cellcolor{myGreen!90}\val%
    \else\ifdim \val pt < -2.0pt \cellcolor{myGreen!80}\val%
    \else\ifdim \val pt < -1.5pt \cellcolor{myGreen!70}\val%
    \else\ifdim \val pt < -1.0pt \cellcolor{myGreen!60}\val%
    \else\ifdim \val pt < -0.7pt \cellcolor{myGreen!50}\val%
    \else\ifdim \val pt < -0.5pt \cellcolor{myGreen!40}\val%
    \else\ifdim \val pt < -0.3pt \cellcolor{myGreen!30}\val%
    \else\ifdim \val pt < -0.2pt \cellcolor{myGreen!25}\val%
    \else\ifdim \val pt < -0.1pt \cellcolor{myGreen!20}\val%
    \else\cellcolor{white}\val%
    \fi\fi\fi\fi\fi\fi\fi\fi\fi\fi
  \else
    \cellcolor{white}\val%
  \fi\fi
  \endgroup
}

\graphicspath{{figs/}{figs/Appendix/}}
\usepackage{xcolor}

\definecolor{MyGreen}{HTML}{708238}

\begin{document}

\maketitle

\begin{abstract}

Mental health research on social media has been dominated by English-language datasets and DSM-5-centered resources, limiting cross-lingual and cross-cultural understanding of online mental health expression. We introduce MATCHA, the first Chinese multi-disorder mental health dataset that links disorder-specific Weibo Super-Topic communities with users' broader public timelines. MATCHA covers seven DSM-5 disorders and a control group, comprising 46,025 cohort-level users and 1.8 million posts. The dataset distinguishes users with explicit self-disclosure evidence from broader disorder-related participants through an LLM-assisted annotation pipeline, and is released in a rigorously de-identified, controlled-access format. Using MATCHA, we analyze behavioral, linguistic, symptom-related, and life-event-related patterns across disorders, and conduct cross-lingual comparisons with Twitter-STMHD. Our analyses show that Weibo Super-Topics function as concentrated spaces for affective and health-related expression, while cross-platform differences between Weibo and Twitter often exceed cross-disorder variation. MATCHA provides a new resource for community-aware and cross-lingual mental health NLP research.
\end{abstract}

\section{Introduction}

Social media platforms are now major channels for expressing emotions, sharing life experiences, and seeking peer support. The resulting user-generated content captures language and behavior continuously, with high temporal resolution and near-real-time availability, providing a complementary evidence stream to surveys and clinical records. In English-language research, resources such as SMHD \cite{cohan-etal-2018-smhd} and Twitter--STMHD \cite{Suhavi:2022:STMHD} enable systematic analyses across multiple disorders and support detection, early identification, and risk modeling.
\begin{figure}[htbp]
    \centering
    \includegraphics[width=\columnwidth]{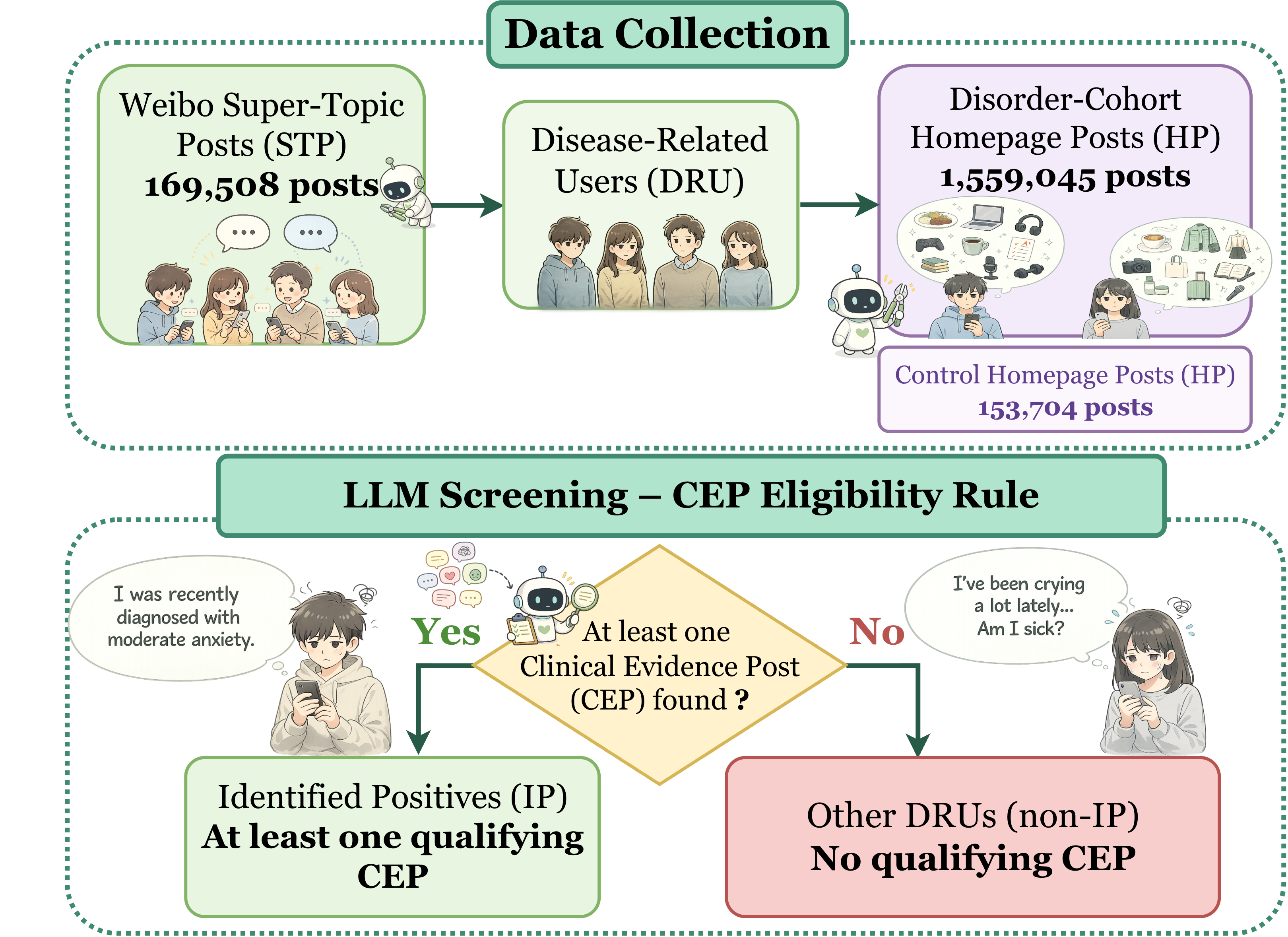}
    \caption{Overview of the MATCHA Dataset Collection and LLM
-assisted Annotation Pipeline.}
    \label{fig:overview}
\end{figure}
Culture and language shape how mental-health experiences are expressed and interpreted \citep{Kleinman1982Neurasthenia}. This motivates mental-health NLP resources that are sensitive to language and community context while remaining explicit about their relationship to major diagnostic taxonomies, such as DSM-5 and ICD-11 \citep{apa2013dsm5,WHO_ICD11_2022}.

Sina Weibo (often referred to as the ``Twitter of China''\footnote{\url{https://weibo.com}}) and its unique Super-Topic mechanism provide an ideal setting to address this research gap. As one of China's largest social platforms, Weibo reported about 591 million monthly active users and 261 million daily active users in the first quarter of 2025 \cite{weibo2025q1}, reflecting its massive influence. Super-Topics are themed sub-communities that aggregate posts around specific interests, including health and mental well-being. This structure enables researchers to jointly analyze community-level discussions and individual-level posting histories, offering a natural context to identify users interested in disorder-related topics via corresponding Super-Topics. As a result, Weibo allows the organic, community-aware collection of mental health content at scale and supports systematic cross-lingual comparisons between Chinese and English-language social media environments.


We introduce MATCHA (Multi-disorder mental heAlth daTa in CHinA), a community-driven Chinese Weibo dataset for multi-disorder mental health research. Covering seven DSM-5 disorders and a control group, it links public user timelines with posts from disorder-specific Super-Topics, followed by de-identification and aggregation. A key feature is its LLM-assisted user stratification into Identified Positives (IP; explicit CEP-defined self-disclosure evidence, including eligible caregiver reports) and Disorder-Related Users (DRU; users discovered through participation in disorder-related Super-Topics, without diagnostic inference). \figref{fig:overview} shows the overview of the MATCHA dataset collection and annotation pipeline. 

Homepage Posts provide a complementary, cue-reduced setting for studying whether cohort-related signals extend beyond disorder-centered communities. Unlike Super-Topic Posts, where disorder-specific terminology and clinical cues are naturally concentrated, Homepage Posts capture users' broader public-timeline language. This design allows us to ask whether user-level cohort-related signals remain detectable in users' broader public timelines, where explicit disorder and clinical cues are substantially less concentrated.

Although MATCHA is a Chinese corpus, its alignment with Twitter-STMHD \citep{Suhavi:2022:STMHD} enables a descriptive cross-platform comparison. MATCHA is aligned with the self-report-oriented, user-level design of Twitter-STMHD while retaining its distinct Weibo Super-Topic candidate-selection and CEP annotation procedure. Twitter-STMHD uses pattern-based anchor tweets for user collection, whereas MATCHA uses disorder-specific Weibo Super-Topics to identify candidate users. 
To this end, our contributions are as follows:

\begin{itemize}[leftmargin=*,noitemsep,topsep=0pt]
\item  We introduce \textbf{MATCHA}, the first multi-disorder Chinese Weibo dataset covering \textbf{7 disorders, over 46k cohort-level users, and 1.8M posts}, linking disorder-specific Super-Topic discourse with users' broader public timelines. This dual-source design supports both community-centered analysis and cue-reduced evaluation beyond disorder-specific communities. We further train DSM-5-grounded \textbf{symptom and life-event classifiers} to support scalable analysis.

\item We establish a \textbf{multi-level analytical framework} that links community dynamics with individual linguistic and behavioral patterns, moving beyond conventional user- or post-level analysis of mental health discourse.

\item We conduct a \textbf{cross-lingual comparison} with Twitter-STMHD, revealing divergent patterns in symptom expression and online behavior, while providing resources for cross-lingual mental health research.
\end{itemize}

\section{Datasets}
\label{sec:datasets}

\begin{figure}[t]
    \centering
    \includegraphics[
        width=\columnwidth,
        trim=5 18 5 5,
        clip
    ]{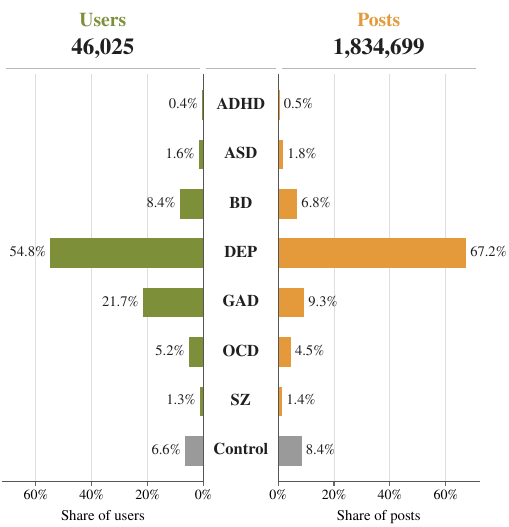}
    \caption{Composition of MATCHA across disorder and control cohorts. Panel (a) reports cohort-level user counts and Panel (b) reports within-cohort deduplicated post counts. Across the eight cohorts, MATCHA contains 46,025 cohort-level users and 1,834,699 within-cohort unique posts. Because cohorts are not mutually exclusive, these aggregated counts may include users or posts appearing in more than one cohort.}
    \label{fig:dataset_overview}
\end{figure}

MATCHA covers seven DSM-5 mental health disorders---attention-deficit/hyperactivity disorder (ADHD), autism spectrum disorder (ASD), bipolar disorder (BD), depressive disorders (DEP), generalized anxiety disorder (GAD), obsessive--compulsive disorder (OCD), and schizophrenia (SZ)---along with a control group. The detailed distribution across diagnostic categories is illustrated in Figure~\ref{fig:dataset_overview}. After deduplication within each cohort, the eight cohorts comprise 46,025 cohort-level users and 1,834,699 unique posts. Because users and posts may occur in multiple disorder cohorts, these aggregated cohort-level counts are not globally deduplicated across cohorts.

\subsection{Data Collection}

Data was collected using Weibo's official APIs under approved research protocols. We identified 1--3 active Super-Topics for each disorder and retrieved posts from two source streams. \textbf{Super-Topic Posts (STP)} are posts retrieved directly from disorder-specific Super-Topic community streams, whereas \textbf{Homepage Posts (HP)} are posts collected from the public personal timelines of users discovered through those communities. Thus, STP and HP are distinguished by retrieval source rather than by manual content categorization. The two sources serve complementary roles. STP captures concentrated disorder-centered discourse, where disorder-specific terminology and clinical cues are naturally more frequent, whereas HP captures users' broader public-timeline language and provides a less cue-saturated setting for examining whether cohort-related signals extend beyond disorder-centered discourse. STP and HP have slightly different temporal coverage, but 99.5\% of disorder-cohort HP fall within the STP date range; detailed collection statistics are provided in Appendix~\ref{app:collection_details}. We also constructed a control group by selecting users discussing neutral topics (e.g., ``breakfast," ``shopping") using 20 everyday keywords.


\subsection{Data Annotation}

Our annotation pipeline consists of two stages: LLM-assisted annotation for high-confidence mental health evidence identification, followed by human annotation for symptom and life-event labeling.

\paragraph{LLM-Assisted Data Annotation.}
Since participation in a Super-Topic does not necessarily imply a clinical diagnosis, we implement a strict \textbf{LLM-assisted annotation protocol} to identify posts containing explicit mental health evidence \cite{Gilardi2023ChatGPTAnnotator}. Specifically, we use a locally deployed \textbf{Qwen3} model \cite{yang2025qwen3} to classify a post as a \textbf{Clinical Evidence Post (CEP)} if it contains one or more of the following signals related to the disorder: (1) an explicit medical diagnosis, (2) mention of prescription medication, (3) a direct self-disclosure or family declaration, or (4) evidence of hospitalization or ongoing treatment. For early-onset disorders such as ADHD and ASD, family-member reports are retained, whereas for other disorders only first-person disclosures are considered. Users with at least one CEP are labeled as \textbf{Identified Positives (IP)}.
To assess validity, we conducted a blinded manual audit of 200 posts: 100 LLM-positive and 100 LLM-negative posts, balanced across the seven disorders. Three researchers independently annotated every post under the CEP guidelines without access to the LLM prediction, and majority vote served as the reference. This audit evaluates agreement with the operational Clinical Evidence Post criterion rather than clinical diagnostic accuracy. Against the majority reference, the LLM achieved 88.0\% precision, 83.0\% recall, and 85.4\% F1. Human reliability was high (87.3\% average pairwise agreement; average pairwise Cohen's $\kappa=0.747$; Fleiss' $\kappa=0.746$; 81.0\% unanimous agreement). On the unanimously labeled subset, LLM precision was 95.8\%. The pipeline exhibits a slightly conservative error profile: its 12 false positives were concentrated in caregiver-oriented or educational/informational posts, especially for ASD, whereas its 18 false negatives more often involved implicit self-reference, weak medication mentions, or ADHD/ASD family-member reports.

\paragraph{Human Annotation.}
Mental-health-related expressions on social media are often implicit and context-dependent, making reliable annotation challenging. To support symptom and life-event annotation, we recruited 21 annotators with backgrounds in psychology, medicine, linguistics, or related disciplines. Annotators received DSM-5-based guideline training and completed pilot annotations before formal labeling.

The team annotated 13,723 unique Weibo posts across 38 symptom and 12 life-event categories following guidelines adapted from prior mental-health studies~\cite{zhang-etal-2022-symptom,lv-etal-2025-tracking}. For posts with multiple independent annotations, labels were aggregated before classifier training. Among multiply annotated posts, macro-averaged Cohen's $\kappa$ across the symptom and life-event labels was 0.72. Detailed annotation and classifier-preparation statistics are provided in Appendix~\ref{app:classifiers}.

\subsection{Dataset Structure}
The final dataset includes:
\begin{itemize}[noitemsep,topsep=0pt]
    \item Super-Topic Posts (STP): Posts from users in disorder-related Weibo Super-Topics.
    \item Homepage Posts (HP): Public personal-timeline posts from users discovered through disorder-related Super-Topics.
    \item Identified Positives (IP): A subset of DRUs with at least one Clinical Evidence Post (CEP), namely explicit textual self-disclosure evidence under the operational CEP criterion, including eligible ADHD/ASD family reports. IP is not a clinically verified diagnosis.
    \item Disorder-Related Users (DRU): Users who posted in disorder-related Super-Topics, irrespective of self-disclosure.
    \item Control Group: Users discussing neutral topics, with posts sampled using everyday keywords.
\end{itemize}


STP and HP therefore serve complementary rather than competing roles. STP captures concentrated, community-centered disorder discourse, whereas HP provides a broader and less cue-saturated public-timeline setting for examining whether cohort-related linguistic and behavioral signals persist beyond disorder-specific communities.

\section{MATCHA Cross-Disorder Analysis}
\label{sec:5.1}

In this section, we present a multi-dimensional exploratory analysis of MATCHA, integrating behavioral, linguistic, and mental-state signals to characterize disorder-related communication on Weibo. We analyze (1) posting behaviors, (2) linguistic features via LIWC, and (3) symptom and life-event signals based on re-trained classifiers. 

\subsection{Posting Behavior}

We first analyze how users interact with the platform across disorders and user types.

\textbf{User Engagement and Posting Volume} Identified Positive (IP) users, though a minority, contribute disproportionately to STP. As shown in Table \ref{tab:ip_activity}, their per-user activity is 1.7--14.7 times higher than non-IP users. This is especially pronounced in ASD, where IPs (under 9\% of STP users) generate nearly 59\% of posts. Across all disorders, HP volumes substantially exceed STP volumes. However, STP/HP ratios vary across disorders, ranging from 0.18 for ASD to 0.06 for schizophrenia, suggesting different levels of engagement with disorder-specific communities.


\begin{table}[t]
\centering
\footnotesize
\setlength{\tabcolsep}{3pt}
\renewcommand{\arraystretch}{1.02}

\caption{Posting activity comparison between CEP-defined IP and non-IP users within the Super-Topic Post (STP) population across disorders. IP denotes explicit textual self-disclosure evidence under the operational CEP criterion, not clinically verified diagnosis.}
\label{tab:ip_activity}

\begin{tabular}{@{}lcccc@{}}
\toprule
Disorder & IP users [\%] & IP posts [\%] & STP/IP & STP/non-IP \\
\midrule
ADHD & 23.5 & 34.5 & 7.6 & 4.4 \\
ASD  & 8.9  & 58.9 & 50.1 & 3.4 \\
BD   & 18.8 & 43.6 & 7.6 & 2.3 \\
DEP  & 19.4 & 50.3 & 12.5 & 3.0 \\
GAD  & 28.6 & 61.7 & 10.3 & 2.6 \\
OCD  & 23.3 & 53.6 & 10.3 & 2.7 \\
SZ   & 18.5 & 38.1 & 5.7 & 2.1 \\
\bottomrule
\end{tabular}
\end{table}

\textbf{Temporal Patterns} All disorder cohorts show higher user-normalized evening posting than the original control cohort (+0.5 to +6.3 percentage points), with the largest increases for DEP, BD, and GAD. Night-time elevation is not uniform: BD, DEP, and OCD have night ratios above the control baseline, whereas the remaining cohorts have lower ratios. Table~\ref{tab:evening-night-metrics} summarizes a user-normalized full-cohort measure; its estimand differs from the pooled-post hourly distributions used for the cross-platform correlation analysis.

\begin{table}[htbp]
\caption{User-normalized evening and night-time posting metrics for the full disorder cohorts, computed on the merged STP+HP corpus after note-ID deduplication. Evening is each user's 20:00--23:00 posting share averaged over users; $\Delta$Evening is relative to the original control cohort; Night Ratio is the 22:00--05:00 share divided by the 10:00--17:00 share.}
\label{tab:evening-night-metrics}
\centering
\small
\setlength{\tabcolsep}{5pt}
\renewcommand{\arraystretch}{1.10}
\begin{tabular}{
l
S[table-format=2.1]
S[table-format=+2.1]
S[table-format=1.2]
}
\toprule
\multicolumn{1}{c}{Disorder} &
\multicolumn{1}{c}{Evening [\%]} &
\multicolumn{1}{c}{$\Delta$Evening [pp]} &
\multicolumn{1}{c}{Night Ratio} \\
\midrule
ADHD & 21.9 & +0.5 & 0.50 \\
ASD  & 22.8 & +1.4 & 0.57 \\
BD   & 27.2 & +5.8 & 1.18 \\
DEP  & 27.7 & +6.3 & 1.20 \\
GAD  & 26.8 & +5.4 & 0.72 \\
OCD  & 26.2 & +4.8 & 0.94 \\
SZ   & 26.5 & +5.1 & 0.79 \\
Control & 21.4 & +0.0 & 0.88 \\
\bottomrule
\end{tabular}
\end{table}

\begin{table*}[!t]
\centering
\setlength{\tabcolsep}{2.0pt}
\renewcommand{\arraystretch}{1.00}
\newcommand{\cc}[1]{\colorcell{#1}}

\caption{
LIWC comparisons between CEP-defined identified-positive (IP) users and final control users on Weibo. IP denotes explicit self-disclosure evidence under the operational Clinical Evidence Post criterion, rather than a clinically verified diagnosis.
\textbf{Panel A} reports results from a Chinese LIWC lexicon applied to the original Chinese posts.
\textbf{Panel B} reports selected categories from the English LIWC applied to LLM-translated posts; all source--disorder translation cohorts achieved exact note-ID coverage.
All values are percentage-point differences in category frequency (\%) between the disorder group and the control group, reported separately for Super-Topic Posts (STP) and Homepage Posts (HP).
}
\label{tab:liwc_cn_en_panels}
\scriptsize
\begin{adjustbox}{width=\textwidth, center}
\begin{tabular}{@{}lrrrrrrrrrrrrrr@{}}
\toprule
\multicolumn{15}{l}{\textbf{Panel A: Chinese LIWC on original Chinese Weibo text}}\\[-2pt]
\midrule
\multirow{2}{*}{\textbf{Cat.}} &
\multicolumn{7}{c}{\textbf{STP}} &
\multicolumn{7}{c}{\textbf{HP}}\\
\cmidrule(lr){2-8}\cmidrule(lr){9-15}
& ADHD & ASD & BD & DEP & GAD & OCD & SZ
& ADHD & ASD & BD & DEP & GAD & OCD & SZ \\
\midrule
\multicolumn{15}{c}{\textit{Affective Processes}}\\
Positive emotion & \cc{-0.31}&\cc{-0.36}&\cc{0.32}&\cc{0.37}&\cc{-0.51}&\cc{-0.39}&\cc{-0.34}&\cc{0.47}&\cc{-0.09}&\cc{0.38}&\cc{0.50}&\cc{0.45}&\cc{0.31}&\cc{0.29}\\
Negative emotion & \cc{0.67}&\cc{0.08}&\cc{2.39}&\cc{2.50}&\cc{2.84}&\cc{3.21}&\cc{2.25}&\cc{0.18}&\cc{0.20}&\cc{0.94}&\cc{1.14}&\cc{0.91}&\cc{0.95}&\cc{0.70}\\
Anxiety & \cc{0.11}&\cc{0.03}&\cc{0.47}&\cc{0.52}&\cc{1.80}&\cc{1.79}&\cc{0.46}&\cc{0.05}&\cc{0.05}&\cc{0.15}&\cc{0.19}&\cc{0.34}&\cc{0.36}&\cc{0.13}\\
Sadness & \cc{0.11}&\cc{0.02}&\cc{0.79}&\cc{0.89}&\cc{0.41}&\cc{0.76}&\cc{0.83}&\cc{0.04}&\cc{0.06}&\cc{0.33}&\cc{0.45}&\cc{0.21}&\cc{0.25}&\cc{0.27}\\
Anger & \cc{0.15}&\cc{-0.04}&\cc{0.42}&\cc{0.40}&\cc{0.16}&\cc{0.31}&\cc{0.36}&\cc{0.02}&\cc{-0.01}&\cc{0.16}&\cc{0.17}&\cc{0.13}&\cc{0.12}&\cc{0.13}\\
\multicolumn{15}{c}{\textit{Cognitive Processes}}\\
Cognitive processes & \cc{7.71}&\cc{3.68}&\cc{10.23}&\cc{10.10}&\cc{10.17}&\cc{12.24}&\cc{8.20}&\cc{3.18}&\cc{3.65}&\cc{3.91}&\cc{3.96}&\cc{3.41}&\cc{4.20}&\cc{1.88}\\
Insight & \cc{2.18}&\cc{1.07}&\cc{3.69}&\cc{3.52}&\cc{3.77}&\cc{4.34}&\cc{2.79}&\cc{1.05}&\cc{0.91}&\cc{1.40}&\cc{1.48}&\cc{1.46}&\cc{1.69}&\cc{0.95}\\
Discrepancy & \cc{1.22}&\cc{0.44}&\cc{1.94}&\cc{1.90}&\cc{1.43}&\cc{2.26}&\cc{1.73}&\cc{0.53}&\cc{0.56}&\cc{0.77}&\cc{0.89}&\cc{0.68}&\cc{0.83}&\cc{0.43}\\
Tentative & \cc{1.31}&\cc{0.48}&\cc{1.97}&\cc{1.85}&\cc{1.56}&\cc{2.25}&\cc{1.82}&\cc{0.41}&\cc{0.56}&\cc{0.77}&\cc{0.82}&\cc{0.63}&\cc{0.82}&\cc{0.37}\\
\multicolumn{15}{c}{\textit{Social Processes}}\\
Social processes & \cc{4.04}&\cc{3.44}&\cc{1.39}&\cc{1.66}&\cc{-0.93}&\cc{0.19}&\cc{1.06}&\cc{1.76}&\cc{2.73}&\cc{0.97}&\cc{1.04}&\cc{0.33}&\cc{0.68}&\cc{1.10}\\
Friends & \cc{-0.11}&\cc{-0.05}&\cc{0.06}&\cc{0.07}&\cc{-0.01}&\cc{-0.00}&\cc{0.04}&\cc{-0.01}&\cc{-0.02}&\cc{0.02}&\cc{0.01}&\cc{0.02}&\cc{0.00}&\cc{0.00}\\
Family & \cc{2.38}&\cc{1.53}&\cc{0.08}&\cc{0.18}&\cc{-0.09}&\cc{-0.13}&\cc{0.32}&\cc{0.68}&\cc{1.21}&\cc{0.09}&\cc{0.07}&\cc{0.08}&\cc{0.06}&\cc{0.24}\\
Humans & \cc{2.83}&\cc{1.91}&\cc{0.84}&\cc{0.76}&\cc{0.19}&\cc{0.61}&\cc{0.45}&\cc{0.66}&\cc{1.58}&\cc{0.41}&\cc{0.39}&\cc{0.19}&\cc{0.32}&\cc{0.30}\\
\midrule
\multicolumn{15}{l}{\textbf{Panel B: EN translation} \quad \textit{Same column order as above: STP(7) $|$ HP(7).}}\\
\midrule
\multicolumn{15}{c}{\textit{Linguistic Dimensions}}\\
I & \cc{1.82}&\cc{0.30}&\cc{7.40}&\cc{7.50}&\cc{6.31}&\cc{6.16}&\cc{5.51}&\cc{1.49}&\cc{0.62}&\cc{3.41}&\cc{3.72}&\cc{3.29}&\cc{3.31}&\cc{2.08}\\
We & \cc{0.23}&\cc{0.37}&\cc{-0.23}&\cc{-0.20}&\cc{-0.32}&\cc{-0.20}&\cc{-0.25}&\cc{0.17}&\cc{0.17}&\cc{-0.03}&\cc{-0.07}&\cc{-0.10}&\cc{-0.06}&\cc{0.04}\\
You & \cc{-0.30}&\cc{-0.63}&\cc{-0.22}&\cc{-0.16}&\cc{-0.44}&\cc{-0.14}&\cc{-0.26}&\cc{0.13}&\cc{-0.24}&\cc{0.24}&\cc{0.46}&\cc{0.16}&\cc{0.23}&\cc{0.17}\\
They & \cc{0.73}&\cc{0.23}&\cc{-0.00}&\cc{0.13}&\cc{-0.06}&\cc{0.15}&\cc{0.18}&\cc{0.11}&\cc{0.35}&\cc{-0.00}&\cc{-0.01}&\cc{-0.05}&\cc{0.00}&\cc{-0.04}\\
She/He & \cc{1.31}&\cc{1.94}&\cc{-0.35}&\cc{-0.26}&\cc{-0.66}&\cc{-0.61}&\cc{-0.15}&\cc{0.81}&\cc{0.74}&\cc{-0.18}&\cc{-0.25}&\cc{-0.30}&\cc{-0.20}&\cc{-0.15}\\
Verbs & \cc{4.80}&\cc{2.32}&\cc{7.79}&\cc{8.24}&\cc{7.00}&\cc{7.11}&\cc{7.06}&\cc{2.93}&\cc{2.27}&\cc{3.62}&\cc{3.85}&\cc{3.64}&\cc{3.78}&\cc{2.11}\\
Adverbs & \cc{1.56}&\cc{0.75}&\cc{3.61}&\cc{4.25}&\cc{3.64}&\cc{3.58}&\cc{2.80}&\cc{1.15}&\cc{0.97}&\cc{1.77}&\cc{1.91}&\cc{2.31}&\cc{2.34}&\cc{0.56}\\
\multicolumn{15}{c}{\textit{Time Orientations}}\\
Past focus & \cc{1.56}&\cc{1.90}&\cc{1.48}&\cc{1.28}&\cc{1.71}&\cc{0.86}&\cc{0.96}&\cc{0.48}&\cc{0.53}&\cc{0.54}&\cc{0.34}&\cc{0.42}&\cc{0.47}&\cc{0.16}\\
Present focus & \cc{2.50}&\cc{0.35}&\cc{5.50}&\cc{6.31}&\cc{4.62}&\cc{5.50}&\cc{5.54}&\cc{2.45}&\cc{1.42}&\cc{2.70}&\cc{3.22}&\cc{2.67}&\cc{2.96}&\cc{1.89}\\
Future focus & \cc{-0.04}&\cc{-0.09}&\cc{0.17}&\cc{0.24}&\cc{0.07}&\cc{0.17}&\cc{-0.08}&\cc{0.05}&\cc{0.04}&\cc{0.13}&\cc{0.15}&\cc{0.11}&\cc{0.12}&\cc{-0.00}\\
\bottomrule
\end{tabular}
\end{adjustbox}
\end{table*}

\subsection{Linguistic Patterns}
\label{sec:liwc}

We conduct LIWC analysis on original Chinese Weibo posts using SC-LIWC~\citep{Gao2013SCLIWC}. English LIWC2015~\citep{Pennebaker2015LIWC} is applied to locally translated versions of the same posts as an auxiliary analysis. Because the Chinese and English analyses use different lexica and languages, we interpret Panel B as an auxiliary analysis rather than a numerically equivalent cross-lingual measurement.



For each LIWC category $c$, we compute its category frequency within a group as:
\begin{equation}
f_c = \frac{N_c}{N_{\text{total}}},
\end{equation}
where $N_c$ is the number of tokens assigned to category $c$, and $N_{\text{total}}$ is the total number of tokens in the corresponding group.
We report the \textbf{percentage-point differences} between CEP-defined Identified Positive (IP) users and controls for both Super-Topic Posts (STP) and Homepage Posts (HP).
Table~\ref{tab:liwc_cn_en_panels} summarizes the results.
Panel A reports Chinese LIWC results on original Chinese Weibo posts, while Panel B reports selected categories from English LIWC computed on translated text.

\textbf{Primary findings from Chinese LIWC.}
Across disorders, we observe consistent affective and cognitive deviations in STP relative to the control group.
Most disorder groups exhibit \textbf{lower positive emotion} and \textbf{higher negative emotion} usage in STPs.
In parallel, \textbf{cognitive process} categories (including \textit{insight}, \textit{discrepancy}, and \textit{tentative}) are substantially elevated across disorders, 
indicating stronger introspection, uncertainty, and self-reflection in disorder-related discussions.

\textbf{Role of English LIWC.}
We additionally apply English LIWC to Chinese posts translated by a locally deployed LLM. English LIWC provides auxiliary estimates for narrative perspective and temporal orientation. It should not be treated as an independent replication of the Chinese LIWC results, because the selected English categories and lexical coverage differ from Panel A.


\textbf{Linguistic dimensions and narrative perspective.}
Panel B shows cross-group differences in pronoun, verb, and adverb categories. ADHD and ASD exhibit relatively stronger third-person pronoun signals (e.g., \textit{she/he}), a pattern compatible with caregiver-mediated narration for early-onset conditions. In contrast, the other disorder cohorts show higher first-person-pronoun use (e.g., \textit{I}) relative to controls, especially in STPs. Verbs and adverbs are also elevated across the reported cohorts. These are descriptive lexical patterns and do not identify the author or speaker of an individual post.

\textbf{Temporal orientation.}
The Chinese LIWC inventory used here does not provide directly comparable tense-focus categories. We therefore report past, present, and future focus only as auxiliary English-LIWC measures. Present-focus is elevated across the reported cohorts, especially in STPs; past-focus is elevated for several cohorts, whereas future-focus differences are small.

\subsection{Exploratory Psychiatric Signal Analysis}

We further analyze symptom-related and life-event-related signals using DSM-aligned bilingual classifiers trained on Weibo and Twitter data. These signals are treated as aggregate proxy indicators rather than clinical diagnoses. Detailed classifier construction and evaluation are provided in Appendix~\ref{app:classifiers}.

To improve cross-lingual consistency and platform-specific robustness, we re-trained separate English and Chinese classifiers for Twitter and Weibo data using updated bilingual annotation procedures. Detailed classifier construction and evaluation results are provided in Appendix~\ref{app:classifiers}.
For all analyses below, \emph{per-post density} for a given signal family $C$ is:
\[
\textstyle \mathrm{Density}(C)=\frac{\sum_{c \in C}\#\text{labels}(c)}{\#\text{posts}}\,,
\]
computed within each disorder $\times$ post type.

\begin{figure}[htbp]
\centering
\captionsetup{font=small,skip=1pt}
\captionsetup[sub]{font=footnotesize,skip=1pt}

\begin{subfigure}[t]{\columnwidth}
  \centering
  \includegraphics[width=1\linewidth]{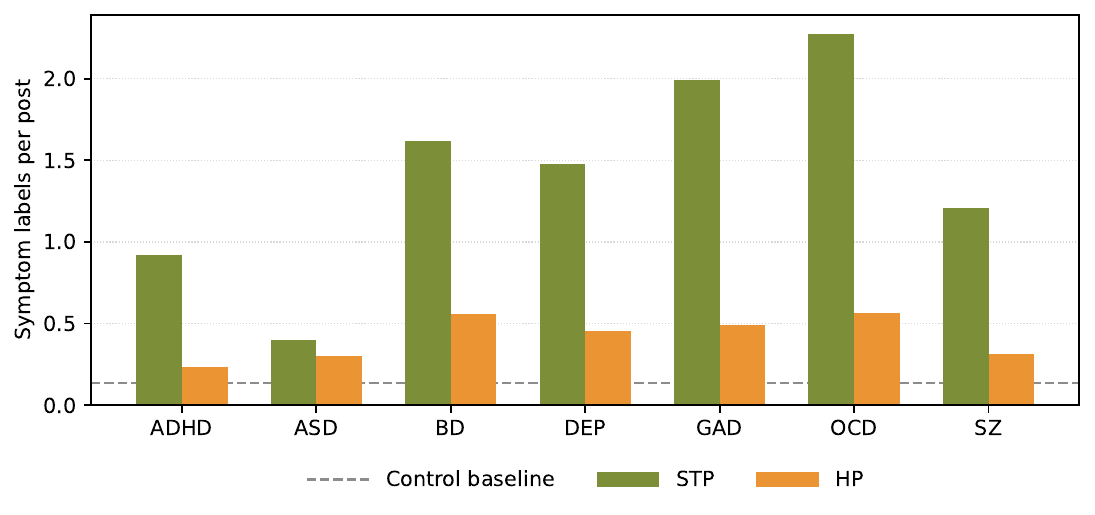}
  \caption{Symptom labels per post}
\end{subfigure}

\vspace{-1pt}

\begin{subfigure}[t]{\columnwidth}
  \centering
  \includegraphics[width=1\linewidth]{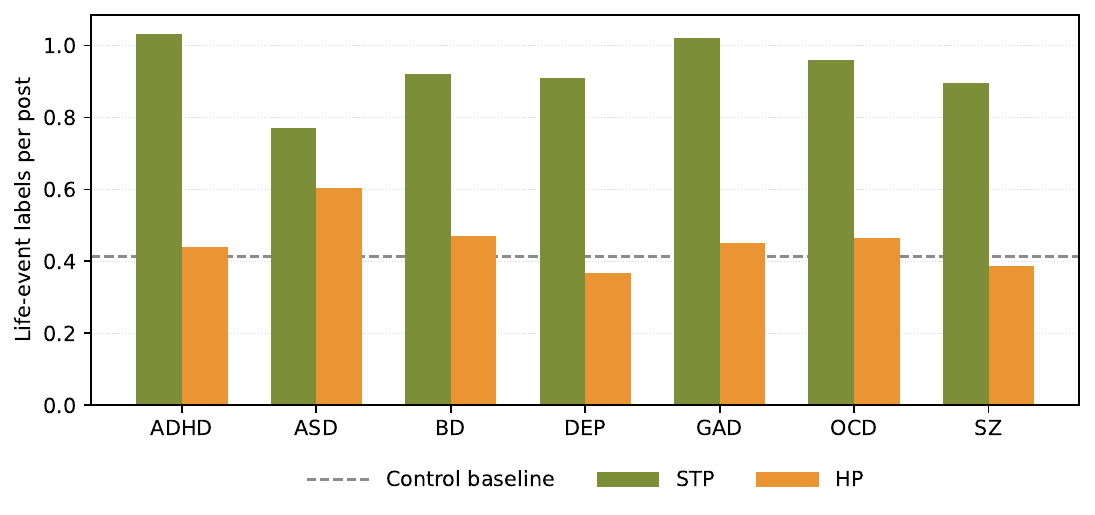}
  \caption{Life-event labels per post}
\end{subfigure}

\caption{Proxy-signal density, defined as the expected number of classifier-positive labels per post, for CEP-defined IP users' STP and HP across disorders. The dashed line denotes the corresponding HP-control baseline. Classifier outputs are aggregate proxies rather than clinical assessments.}
\label{fig:matcha_density_duo}
\vspace{-3pt}
\end{figure}

\textbf{Context Amplifies Mental Health Signals}
Figure~\ref{fig:matcha_density_duo} shows that STP densities exceed HP densities for both signal families in every disorder, with the magnitude of the difference varying by disorder and signal family. Super-topic Posts (STPs) show higher densities for both symptoms and life events, suggesting that STPs provide a more information-dense context for mental-health-related discourse than HPs.
\textbf{Disorder-Specific Content Profiles}
The magnitude of the STP--HP gap varies by disorder and signal family. Symptom density is highest in OCD and GAD STP, whereas life-event density is relatively high in ADHD and GAD STP. We interpret these classifier-derived differences descriptively; they do not identify individual clinical status or establish the source of a cohort-level pattern.

\subsection{Cue-Reduced Evaluation with Homepage Posts}
\label{sec:cue_reduced_hp}

We further examine whether cohort-related signals persist beyond disorder-centered communities. HP-only achieves an average Macro-F1 of 0.842 (AUROC 0.932), compared with 0.953 (0.995) for STP-only, while explicit disorder or clinical cues occur substantially less often in HP than STP (4.64\% vs.\ 19.74\%). After masking these cues, Macro-F1 decreases by 0.009 for HP and 0.017 for STP; the average reduction is larger for STP. These results position HP as a complementary cue-reduced setting in which cohort-related signals remain detectable beyond disorder-specific communities. This benchmark predicts cohort membership rather than clinical diagnosis. Full experimental settings and per-disorder results are provided in Appendix~\ref{app:hp_cue_reduced}.

Across analyses, a consistent regularity is \emph{context amplification}: STP concentrate substantially more classifier-derived proxy signal than HP, although the magnitude of this difference varies by disorder and proxy-signal family. This pattern supports the interpretation that STP offer an information-dense window into mental-health-related discourse, whereas HP reflects broader everyday public-timeline language.

\section{Cross-lingual Analysis}

We compare mental health expression between Weibo (MATCHA) and Twitter-STMHD\cite{Suhavi:2022:STMHD} across shared disorder categories. To ensure balanced comparison, we restrict the comparison to BD, DEP, GAD, and OCD, the four cohorts satisfying the shared-data and label-mapping requirements. ADHD, ASD, and SZ are not included in these cross-platform figures.

\subsection{Linguistic Pattern Comparison}

\begin{figure*}[htbp] 
    \centering
   \includegraphics[width=\textwidth]{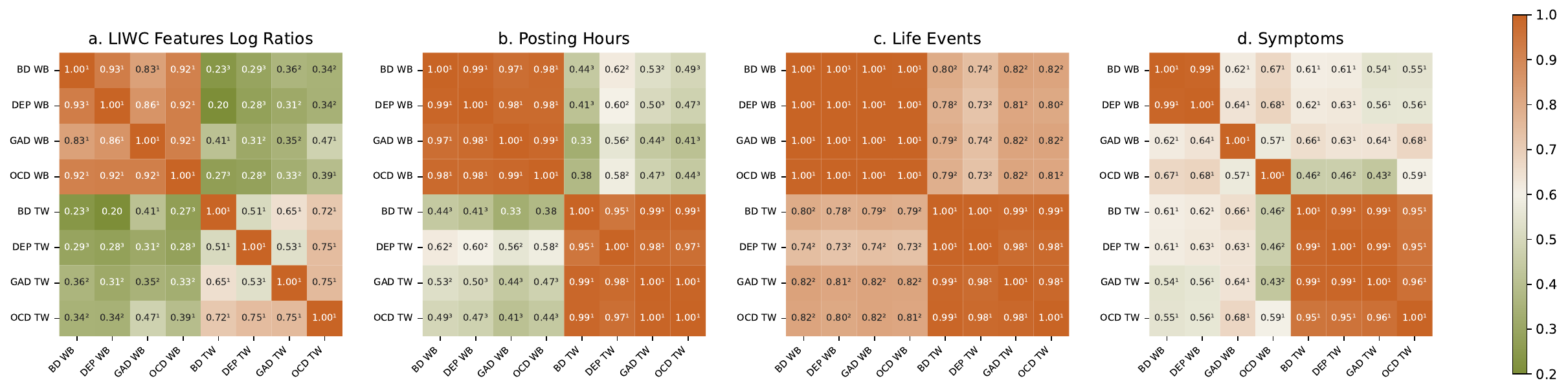}
    \caption{
    Cross-lingual Pearson correlation matrices for (a) LIWC feature log ratios, (b) posting hours, (c) life-event distributions, and (d) symptom distributions. Cross-platform symptom correlations in (d) are lower on average than within-platform correlations, indicating substantial platform- and language-associated variation in symptom composition. These correlations are descriptive and do not identify causal cultural effects. Numbers in the upper-right corner of each cell denote statistical significance levels: 1 = $p < 0.001$, 2 = $p < 0.01$, and 3 = $p < 0.05$.}
  \label{fig:correlation_matrix}
\end{figure*}

We extracted LIWC word frequencies from Twitter-STMHD using the same procedure as for MATCHA. To align with the self-reported STMHD setting, we use only CEP-defined Identified Positive users from MATCHA and merge each user's Super-Topic and homepage posts. For each LIWC category $c$, we compute a cohort-control log ratio within each platform:
\begin{equation}
\label{equa:2}
\text{LR}(c) = \log \frac{f_{\text{disorder},c} + \varepsilon}{f_{\text{control},c} + \varepsilon}
\end{equation}
where $\varepsilon = 10^{-6}$. Correlations over the 73 LIWC categories show stronger within-platform than cross-platform similarity: disorder groups are closer to other groups in the same language than to their disorder counterparts across platforms (Figure~\ref{fig:correlation_matrix}a). This pattern suggests that platform- and language-associated variation can exceed disorder-specific variation in LIWC signals.

\begin{figure}[hbpt]
  \centering
  \includegraphics[width=1\columnwidth]{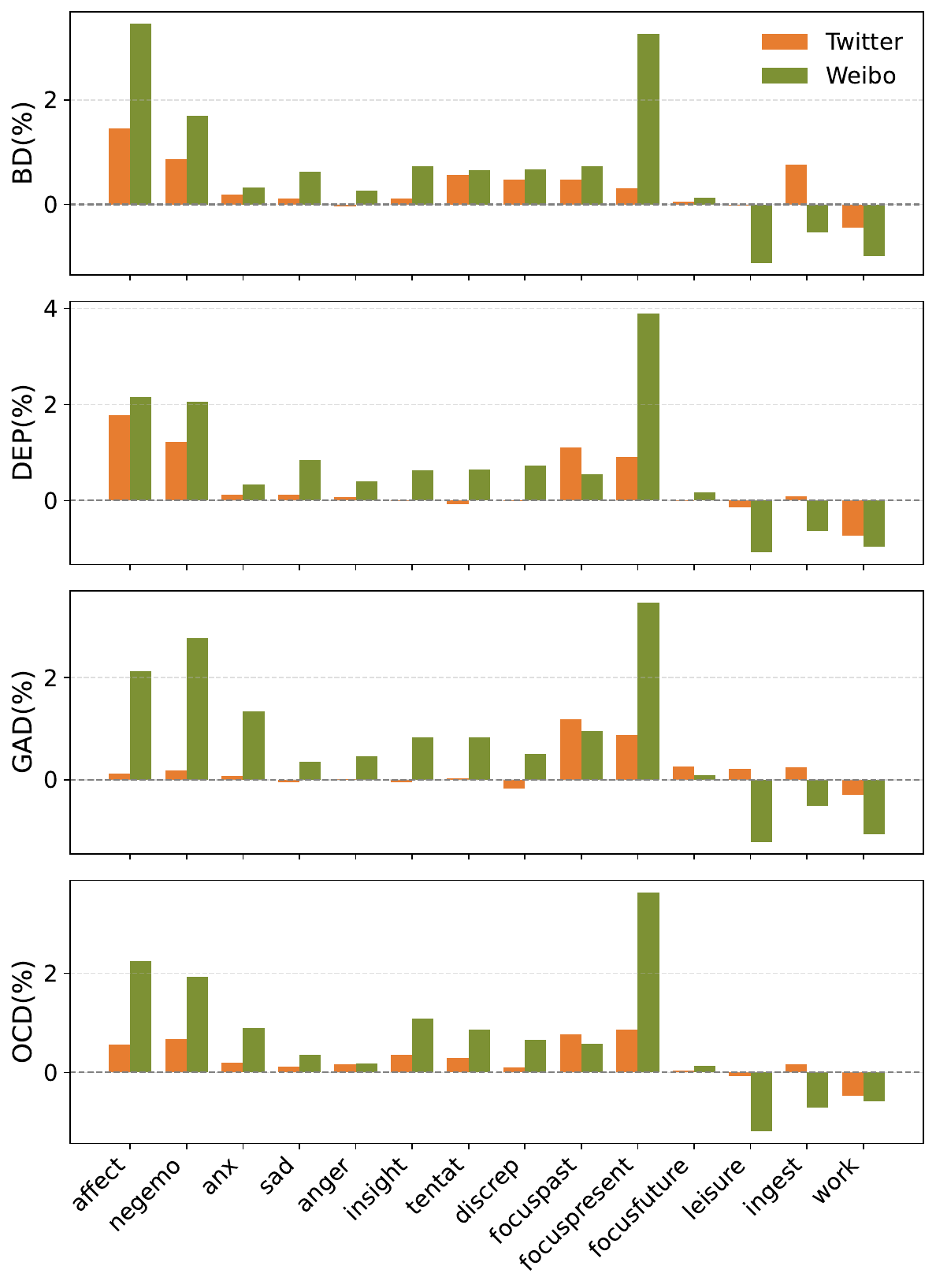}
  \caption{Cross-platform differences in core LIWC features across BD, DEP, GAD, and OCD. Bars show within-platform cohort--control frequency differences (percentage points) for CEP-defined MATCHA IP cohorts and frozen Twitter-STMHD cohorts.}
  \label{fig:liwccomparaison}
\end{figure}

We further inspect representative LIWC categories with large cohort-control deviations (Figure~\ref{fig:liwccomparaison}). Both platforms exhibit elevated affective expression, but MATCHA shows stronger increases in negative emotion, anxiety, sadness, and cognitive-processing categories. MATCHA users also display stronger present-focus patterns, whereas Twitter-STMHD exhibits relatively greater past-focus. In addition, life-related categories such as leisure and ingest are lower in MATCHA than in Twitter-STMHD. These comparisons are descriptive within-platform cohort--control contrasts rather than direct raw-frequency comparisons.

\subsection{Disorder Presentation Comparison}

We next compare normalized posting-hour, symptom, and life-event distributions. The Chinese and English classifiers are trained separately with bilingual annotations, so all results are interpreted as aggregate proxy trends rather than individual-level clinical claims.


Across these three views, groups cluster primarily by platform/language rather than disorder type. Within-platform correlations are generally higher than cross-platform correlations. This structure indicates substantial platform- and language-associated variation in online disorder presentation, potentially reflecting the combined effects of platform affordances, community structure, sampling, and cultural context.


Figure~\ref{fig:correlation_matrix}c summarizes the life-event correlation pattern. As with the other cross-platform measures, these correlations are descriptive and may reflect differences in sampling, label mapping, classifier behavior, language, and platform context.

Because classifier design, platform sampling, and label mapping differ across platforms, these symptom-composition comparisons are descriptive and should not be interpreted as evidence for clinical or causal cross-cultural mechanisms.

\begin{figure}[htbp]
    \centering
    \includegraphics[width=\columnwidth]{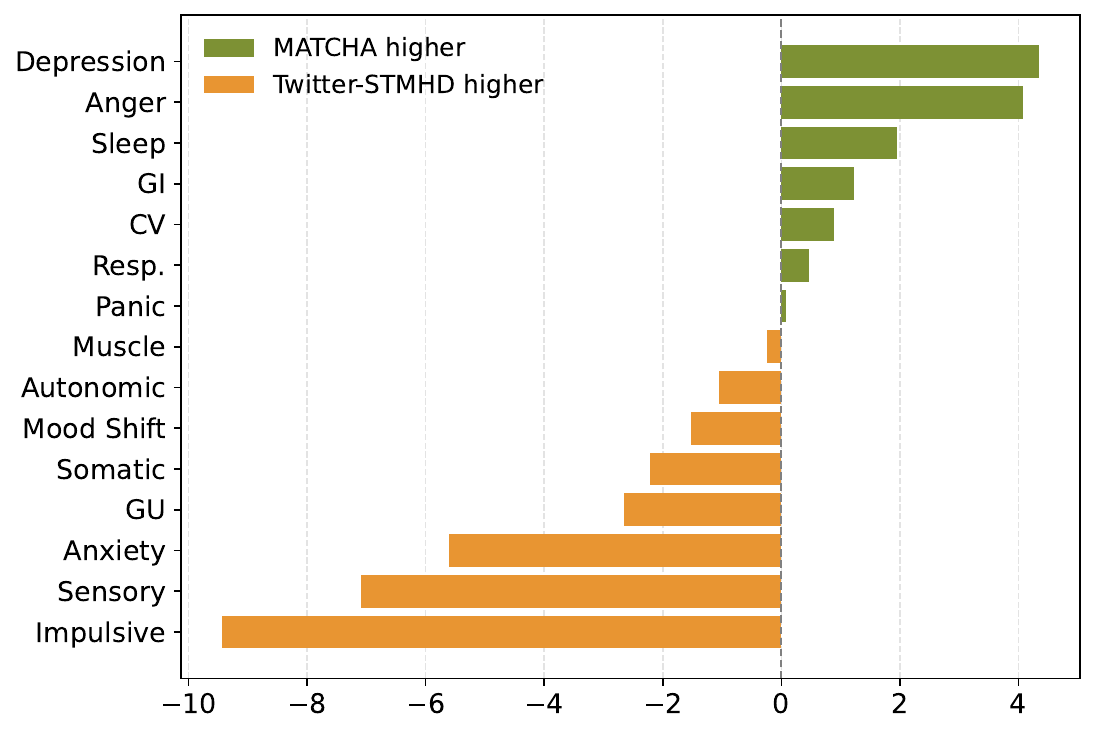}
    \caption{Cross-platform symptom-category distribution differences between MATCHA and Twitter-STMHD. Avoidance is excluded from STMHD due to unstable English classifier behavior.}
    \label{fig:overall_symp_distribution_difference}
\end{figure}

Figure~\ref{fig:overall_symp_distribution_difference} summarizes the average categorywise symptom-prevalence differences (MATCHA minus Twitter--STMHD) across the four shared disorder cohorts. Avoidance is omitted because its English classifier estimates were unstable.

\paragraph{Summary}
Overall, the cross-platform analyses reveal larger between-platform/language differences than cross-disorder differences across several measured dimensions, but these patterns should be interpreted descriptively because the present design cannot isolate cultural effects from platform, language, sampling, demographic, and community-level factors.

\section{Related Work}

\textbf{Self-Report-Based Mental Health Social Media Datasets}
Foundational datasets such as SMHD \cite{cohan-etal-2018-smhd} and Twitter-STMHD \citep{Suhavi:2022:STMHD} link social media users to DSM-5 disorders through explicit self-reporting. SMHD identifies diagnosed Reddit users via high-precision patterns while removing mental-health-related content to reduce topical bias, whereas STMHD extends this framework to Twitter with multi-year timelines across eight disorders. However, both primarily rely on explicit self-disclosure and may overlook users expressing symptoms indirectly, particularly in high-stigma contexts. Our work preserves their DSM-5 and user-level framework while incorporating disorder-related community participants identified through Super-Topic participation, with a separate LLM-assisted procedure used to identify explicit self-disclosure evidence. Instead of de-topicalization, we leverage Weibo Super Topics as organic community spaces for analyzing support and disorder-related discourse.

\noindent\textbf{Chinese Social Media Mental Health Corpora}
In the Chinese setting, WU3D \citep{Wang:2022:WU3D} provides a professionally annotated Weibo dataset for depression detection and serves as a benchmark for supervised classification. However, WU3D focuses on depression and does not provide MATCHA's dual STP/HP retrieval design or its operational CEP-based IP definition. MATCHA instead covers multiple disorder cohorts and supports descriptive cross-platform comparison with English social-media data.

\noindent\textbf{Early Detection and Shared Task Initiatives}
Resources such as RSDD \citep{Yates:2017:RSDD}, CLPsych 2015 \citep{Coppersmith:2015:CLPsych}, and CLEF eRisk \citep{Losada:2017:eRisk} established benchmarks for mental-health detection and early-risk prediction in English social media. These resources use platform-specific self-report or clinician-provided proxy labels; their collection and topical-filtering choices differ by resource. In contrast, we study disorder-related community participants on Chinese social media without treating community participation as evidence of diagnosis.

\section{Conclusion}

We present MATCHA, a large-scale Chinese mental health social media dataset covering seven DSM-5 disorders and a control group. By linking Weibo Super-Topic communities with user-level timelines, MATCHA provides an ethically de-identified corpus that connects clinical concepts with real-world mental health discourse. Our analyses show that Super-Topics function as high-density contexts for disorder-related expression, featuring elevated negative affect, richer life-event and symptom signals, and higher user activity. Cross-lingual comparison further reveals that platform- and language-level differences between Weibo and Twitter are more pronounced than cross-disorder variations, highlighting substantial variation associated with linguistic and platform environments in online mental health expression. Some observed differences may also reflect classifier transfer effects and residual cross-lingual annotation mismatches. 

Our cue-reduced evaluation further shows that Homepage Posts retain cohort-related signals despite containing substantially fewer explicit disorder cues than Super-Topic Posts, highlighting the complementary roles of community-centered and broader timeline discourse. Together, MATCHA supports descriptive analysis, cross-platform comparison, and downstream cohort-level modeling of mental-health-related social-media language.

\section*{Limitations}

Several limitations should be considered when interpreting MATCHA and the analyses presented in this work.

First, MATCHA is constructed from public Weibo Super-Topic communities and therefore does not represent the broader population of individuals with mental health conditions in China. Users who actively participate in disorder-related communities are likely more willing to disclose symptoms, seek support, or engage in mental-health-related discussion, which may bias linguistic and behavioral patterns toward highly engaged online populations.

Second, although the dataset is organized around DSM-5 disorder categories, online self-disclosure and community participation do not constitute clinical diagnosis. Our Identified Positive (IP) labels rely on explicit textual evidence of diagnosis, treatment, self-disclosure, or eligible family reports, rather than verified medical records, and the symptom/life-event analyses are based on classifier-derived proxy signals intended only for aggregate-level comparisons. The cue-reduced benchmark likewise predicts cohort membership rather than clinical diagnosis and should not be interpreted as inferring mental disorders from ordinary social-media posts.

Third, cross-lingual comparisons between Weibo and Twitter-STMHD remain affected by substantial cultural, linguistic, and platform-specific differences. Variations in moderation policies, posting norms, demographic composition, and discourse conventions may contribute to observed differences independently of mental-health status. In addition, LIWC categories are not perfectly aligned across Chinese and English, even when translation-based robustness checks are applied.

Fourth, the dataset currently focuses on seven DSM-5 disorders and does not cover the full spectrum of psychiatric or neurodevelopmental conditions. Comorbidity, symptom overlap, and temporal disorder progression are also not explicitly modeled in the current release.

Finally, despite rigorous de-identification and controlled-access release procedures, mental-health social-media data inherently involve privacy and ethical risks. Future work should continue exploring stronger privacy-preserving release mechanisms, culturally adaptive annotation frameworks, and clinically informed evaluation protocols.

\section*{Ethical Considerations}

All data were collected from publicly accessible posts on Weibo through the platform's official APIs under an approved research application, in accordance with Weibo's terms of service. No private, deleted, or restricted content was accessed. The data collection process was non-intrusive and did not interfere with normal user activities. Individual informed consent was not sought from Weibo users because this study used only publicly accessible posts retrieved through the platform's official APIs under an approved research application. To mitigate privacy risks, we de-identified the data and release only controlled-access, aggregate statistics and derived features.

Upon publication, the MATCHA dataset is available through a controlled-access mechanism for academic, non-commercial research only. The released data will not contain raw user identifiers or direct personal information; instead, they will consist of de-identified, aggregated statistics and derived features. The accompanying license will explicitly prohibit any attempt at user re-identification or deanonymization. Documentation, the Data Use Agreement, and controlled-access request instructions are available at
\url{https://github.com/Baihan-12/MATCHA-Mental-Health-Dataset}.

To protect user privacy, all data underwent strict de-identification, including the removal or hashing of usernames, user IDs, and URLs. For cross-lingual analysis, Chinese posts were translated solely for analytical purposes, which additionally reduces the risk of identifying individual writing styles. All processing was conducted using locally deployed large language models in offline environments, with no external data transmission, further minimizing privacy and security risks.

Potential risks include privacy leakage, re-identification attempts, stigmatization, and inappropriate use of cohort-level or classifier-derived signals for individual-level clinical inference. We mitigate these risks through controlled access, de-identification, explicit prohibitions on re-identification, and by restricting all reported analyses to aggregate-level research purposes.

\section*{Acknowledgements} This work has been supported by the China NSFC Project (No. 62572320).

\bibliography{ijcai26}

\newpage
\appendix

\setcounter{topnumber}{3}
\setcounter{bottomnumber}{2}
\setcounter{totalnumber}{5}

\setcounter{dbltopnumber}{2}
\renewcommand{\dbltopfraction}{0.95}
\renewcommand{\dblfloatpagefraction}{0.9}

\section{Additional Dataset Collection Details}
\label{app:collection_details}

\paragraph{Temporal coverage.}
The main STP collection spans May 13, 2014--January 14, 2025. Disorder-cohort HP extend further back in time, with the earliest retrieved post dated April 10, 2010, because homepage retrieval provides historical public-timeline context for users identified through Super-Topic communities. Control HP span October 17, 2009--December 29, 2024.

Despite this broader nominal range, the temporal mismatch is small in practice: 1,551,872 of 1,559,045 disorder-cohort HP (99.5\%) fall within the STP date range. Of the 7,173 posts outside this range, 7,070 precede May 13, 2014, reflecting the longer historical reach of homepage retrieval, while only 103 occur after January 14, 2025. The latter comprise 103 additional GAD homepage posts from 10 users,
extending to September 30, 2025, without corresponding additional
STP collection. Restricting HP to the STP date range retains 30,843 of 30,851 users ($>$99.9\%). Thus, nearly all HP data overlap temporally with STP, making it unlikely that the broader nominal HP range alone accounts for the observed differences between the two sources.

\paragraph{Homepage-post retrieval.}
The released data are consistent with homepage retrieval of up to approximately 100 recent public-timeline posts per user at a retrieval point. Some users have more than 100 unique HP in the final corpus, consistent with repeated retrieval and subsequent post-ID deduplication. Consequently, the final number of unique HP per user is not strictly capped at 100. Across disorder cohorts, users have a median of 36 HP (IQR: 5--96; mean: 50.5); 10.0\% have exactly 100 posts and 4.0\% have more than 100. The earliest retrieved HP precedes a user's first observed STP by a median of 222 days (IQR: 37--714 days), reflecting the broader historical timeline context captured by HP.

\paragraph{Super-Topic search.}
For each disorder, we queried one to three public Weibo Super-Topics
using common Chinese disorder names and terminology variants.
Table~\ref{tab:topic_search_verification} records the disorder cohorts
and their corresponding Super-Topic terms. It documents the initial
collection configuration; final corpus counts arise only after
downstream user expansion and post-ID deduplication and are therefore
not expected to equal the final corpus counts reported in the main
paper. The Super-Topic mapping identifies public discussion communities
and does not establish that any participant has a clinical diagnosis.

\begin{table*}[t]
\centering
\small
\setlength{\tabcolsep}{4pt}
\renewcommand{\arraystretch}{1.06}
\caption{Disorder cohorts and their corresponding public Weibo
Super-Topics. Chinese terms are accompanied by pinyin transliterations
and English glosses.}
\label{tab:topic_search_verification}
\begin{tabularx}{\textwidth}{@{}l>{\raggedright\arraybackslash}X@{}}
\toprule
Cohort & Weibo Super-Topic \\
\midrule
DEP  & \zh{抑郁症超话} (\textit{yiyuzheng chaohua}, `Depression Super-Topic') \\
DEP  & \zh{抑郁症患者超话} (\textit{yiyuzheng huanzhe chaohua}, `Depression Patient Super-Topic') \\
DEP  & \zh{抑郁超话} (\textit{yiyu chaohua}, `Depression Super-Topic') \\
SZ   & \zh{精神分裂症超话} (\textit{jingshen fenliezheng chaohua}, `Schizophrenia Super-Topic') \\
GAD  & \zh{焦虑症超话} (\textit{jiaolvzheng chaohua}, `Anxiety Disorder Super-Topic') \\
ADHD & \zh{多动症超话} (\textit{duodongzheng chaohua}, `Hyperactivity Disorder Super-Topic') \\
ASD  & \zh{自闭症超话} (\textit{zibizheng chaohua}, `Autism Super-Topic') \\
BD   & \zh{躁郁症超话} (\textit{zaoyuzheng chaohua}, `Manic-Depressive Disorder Super-Topic') \\
BD   & \zh{双相情感障碍超话} (\textit{shuangxiang qinggan zhangai chaohua}, `Bipolar Affective Disorder Super-Topic') \\
OCD  & \zh{强迫症超话} (\textit{qiangpozheng chaohua}, `Obsessive-Compulsive Disorder Super-Topic') \\
\bottomrule
\end{tabularx}
\end{table*}

\paragraph{Control anchor search.}
The control cohort was constructed through a separate everyday-topic
search pipeline rather than through a Super-Topic. We used 20
everyday-life keywords: \zh{上班} (\textit{shangban}, `going to work'),
\zh{早餐} (\textit{zaocan}, `breakfast'),
\zh{聊天} (\textit{liaotian}, `chatting'),
\zh{购物} (\textit{gouwu}, `shopping'),
\zh{健身} (\textit{jianshen}, `fitness'),
\zh{音乐} (\textit{yinyue}, `music'),
\zh{旅行} (\textit{lvxing}, `travel'),
\zh{逛街} (\textit{guangjie}, `shopping or strolling'),
\zh{火锅} (\textit{huoguo}, `hot pot'),
\zh{小吃} (\textit{xiaochi}, `snacks'),
\zh{生日} (\textit{shengri}, `birthday'),
\zh{假期} (\textit{jiaqi}, `holiday'),
\zh{手机} (\textit{shouji}, `mobile phone'),
\zh{游戏} (\textit{youxi}, `games'),
\zh{电话} (\textit{dianhua}, `telephone'),
\zh{分享} (\textit{fenxiang}, `sharing'),
\zh{天气} (\textit{tianqi}, `weather'),
\zh{晚上} (\textit{wanshang}, `evening'),
\zh{公交} (\textit{gongjiao}, `public bus'), and
\zh{路上} (\textit{lushang}, `on the road').
This search produced 3,992 raw anchor posts before downstream user
expansion and deduplication.

\section{Life Event and Symptom Classifiers}
\label{app:classifiers}

To support cross-lingual psychiatric signal analysis, we constructed separate English and Chinese classifiers for symptom-related and life-event-related categories. Our approach builds upon prior DSM-aligned psychiatric signal frameworks, including PsySym for symptom categories and prior work on psychiatric life-event analysis \citep{chen-etal-2024-mapping, lv-etal-2025-tracking, zhang-etal-2022-symptom}.

\paragraph{Annotation.}
The merged annotation corpus contains 13,723 unique posts labeled by a team of 21 annotators across 38 symptom and 12 life-event categories. For each label, annotators assigned 1 if the post mentioned the category and the author was currently experiencing it or still affected by it, 0 if the category was mentioned but was not clearly attributable to the author's current experience, and $-1$ if the category was not mentioned. For posts with multiple independent annotations, labels were aggregated before classifier training. All annotators were fairly compensated at or above the local living wage, with rates determined prior to annotation.

\paragraph{Annotator recruitment and compensation.}
We recruited 21 annotators through social-media recruitment. All annotators were based in China and had backgrounds in psychology, medicine, linguistics, or related fields. Annotators received a total of RMB~13{,}179.79 for 22{,}153 completed annotation records, corresponding to RMB~0.595 per record. The expected annotation time was approximately one minute per record, yielding an estimated compensation rate of RMB~35.70 per hour. Rates were determined before annotation and were at or above the applicable local living-wage benchmark.

\paragraph{Inter-annotator agreement.}
Inter-annotator agreement was assessed on the 6,612 posts with at
least two independent annotators and valid 50-dimensional label
vectors. Cohen's $\kappa$ was computed per label from pairwise
annotator comparisons, including only instances for which both
annotators judged the category to be mentioned (0 or 1); cells coded
as not mentioned ($-1$) were excluded. One label lacked sufficient
variance in this subset and was excluded. The macro-average over the
49 evaluable labels was $\kappa=0.72$.

\paragraph{Data preparation.}
Since our classifiers target relevance, i.e., whether a symptom or life event is mentioned in a post, both 0 and 1 were treated as positive votes during majority-voting aggregation across annotators, while $-1$ was treated as a negative vote. The aggregated label therefore reflects whether annotators reached majority consensus on relevance.

For symptom classification, we applied a disease mask: on symptom dimensions atypical for a post's disorder, aggregated 0 labels were set to $-1$ and excluded from loss computation, avoiding penalization for plausible false negatives. For life-event classification, no disease mask was applied. Instead, $-1$ labels were treated as negative votes during aggregation, such that cells where the majority vote was $-1$ received a final label of 0 (no event).

To mitigate class imbalance, we applied a negative downsampling ratio of 10:1 per label column, masking excess negative samples to $-1$ and excluding them from loss computation. Posts were split into train, validation, and test sets at 70\%/10\%/20\% using iterative stratified splitting, ensuring sufficient positive examples for rare labels in both validation and test sets.

\paragraph{Classifier training.}
We conducted grid search over two model families for each task.

For symptom classification:
\begin{itemize}
    \item \textbf{Chinese-MentalBERT} ($\sim$102M trainable parameters):
    learning rate $\in \{1\times10^{-4}, 2\times10^{-4}\}$,
    batch size $\in \{16,32\}$,
    max length $\in \{64,128,192\}$.

    \item \textbf{Qwen3-Embedding-8B} (frozen backbone, $\sim$49.2K trainable parameters):
    learning rate $\in \{5\times10^{-5},1\times10^{-4},2\times10^{-4}\}$,
    batch size $\in \{8,16\}$,
    max length $\in \{256,512\}$.
\end{itemize}

Chinese-MentalBERT was selected based on validation macro-AUC. The final symptom checkpoint used learning rate $1\times10^{-4}$, batch size 32, and maximum length 128, and achieved validation macro-AUC 0.9502.

For life-event classification, Qwen3-Embedding-8B was selected based on validation macro-AUC. The final checkpoint used learning rate $5\times10^{-5}$, batch size 8, and maximum length 256, and achieved validation macro-AUC 0.9542.

In both tasks, $-1$ labels were masked from the binary cross-entropy loss during training. Positive class weights were set to $\sqrt{n_{\mathrm{neg}}/n_{\mathrm{pos}}}$ per label to further mitigate imbalance. Final checkpoints were selected based on validation macro-AUC and re-evaluated on held-out test sets.

\paragraph{English classifiers.}
For cross-lingual comparison, we re-trained English classifiers on Twitter data to mitigate potential domain shift from the original Reddit-based models. Candidate annotations were obtained by applying existing English symptom and life-event classifiers to Twitter-STMHD, followed by verification using Qwen3-8B. The resulting dataset was matched in scale to the Chinese annotation set and used to train updated English classifiers based on \texttt{mental-bert-base-uncased}, following the same training procedure as the Chinese classifiers.

\paragraph{Annotation statistics.}
Table~\ref{tab:symptom_label_stats} and Table~\ref{tab:lifeevent_label_stats} provide detailed per-label annotation statistics for the Chinese symptom and life-event datasets, respectively. We report both the number of strictly positive annotations (label value 1) and the total number of relevance annotations (0+1) prior to majority-vote aggregation. Because the classification objective focuses on whether a symptom or life event is mentioned in a post, both 0 and 1 are treated as relevance-positive signals during aggregation, while $-1$ denotes absence of mention. The tables further illustrate the substantial class imbalance present across both tasks, motivating the use of disease masking, negative downsampling, and class-balanced loss weighting during classifier training.

\paragraph{Classifier usage.}
The resulting classifiers were used only for aggregate-level proxy signal analysis in the main paper. All symptom- and life-event-related results should therefore be interpreted as distributional trends rather than clinical assessments or individual-level diagnoses.

\begin{table}[t]
\centering
\footnotesize
\setlength{\tabcolsep}{3pt}
\caption{Annotation statistics for the Chinese training data. The 22,153 records are shared post--annotator records annotated for both task families; the two columns are not additive.}
\label{tab:annotation_stats}

\begin{tabular}{lcc}
\toprule
 & Symptom & Life Event \\
\midrule
Number of labels & 38 & 12 \\
Annotated posts & 13,723 & 13,723 \\
Annotation records & 22,153 & 22,153 \\
Median positive rate & 2.26\% & 1.47\% \\
Positive rate range & 0.78--16.39\% & 0.26--43.94\% \\
Most frequent label & Depressed Mood & Health \\
Most frequent rate & 16.39\% & 43.94\% \\
Least frequent label & Impulsivity & Legal \\
Disease mask & Yes & No \\
Negative downsampling & 10:1 & 10:1 \\
Train / Val / Test & 70/10/20 & 70/10/20 \\
\bottomrule
\end{tabular}

\end{table}

\begin{table}[t]
\centering
\small
\caption{Chinese classifier evaluation on a held-out test set ($n=2{,}745$). Per-label decision thresholds were selected on the validation set and applied once to the held-out test set.}
\label{tab:classifier_results_zh}
\begin{tabular}{lccc}
\toprule
Task & Macro AUC & Micro AUC & Macro F1 \\
\midrule
Symptom    & 0.9374 & 0.9494 & 0.5579 \\
Life Event & 0.9533 & 0.9673 & 0.6449 \\
\bottomrule
\end{tabular}
\end{table}

\begin{table}[t]
\centering
\small
\caption{English classifier evaluation results on held-out test sets.}
\label{tab:classifier_results_en}
\begin{tabular}{lccc}
\toprule
Task & Macro AUC & Micro AUC & Macro F1 \\
\midrule
Symptom & 0.9910 & 0.9942 & 0.8852 \\
Life Event & 0.9769 & 0.9767 & 0.7669 \\
\bottomrule
\end{tabular}
\end{table}

\begin{table*}[t]
\centering
\small
\caption{Per-label annotation statistics for symptom labels. Positive counts correspond to value 1, while mention counts include both 0 and 1, i.e., instances in which the corresponding category was mentioned in the post.}
\label{tab:symptom_label_stats}
\begin{tabular}{clrrrr}
\toprule
\# & Label & Positive (1) & Positive \% & Mentioned (0+1) & Mentioned \% \\
\midrule
1 & Irritable Mood & 372 & 2.71 & 499 & 3.64 \\
2 & Anxious Mood & 1,498 & 10.92 & 1,835 & 13.37 \\
3 & Autonomic Symptoms & 834 & 6.08 & 899 & 6.55 \\
4 & Cardiovascular Symptoms & 740 & 5.39 & 786 & 5.73 \\
5 & Catatonic Behavior & 197 & 1.44 & 244 & 1.78 \\
6 & Fatigue or Loss of Energy & 550 & 4.01 & 622 & 4.53 \\
7 & Depressed Mood & 2,249 & 16.39 & 2,645 & 19.27 \\
8 & Gastrointestinal Symptoms & 531 & 3.87 & 591 & 4.31 \\
9 & Genitourinary Symptoms & 134 & 0.98 & 185 & 1.35 \\
10 & Hyperactivity / Agitation & 176 & 1.28 & 250 & 1.82 \\
11 & Impulsivity & 107 & 0.78 & 282 & 2.05 \\
12 & Inattention & 248 & 1.81 & 355 & 2.59 \\
13 & Indecisiveness & 128 & 0.93 & 151 & 1.10 \\
14 & Respiratory Symptoms & 657 & 4.79 & 703 & 5.12 \\
15 & Suicidal Ideation & 783 & 5.71 & 899 & 6.55 \\
16 & Worthlessness / Guilt & 642 & 4.68 & 756 & 5.51 \\
17 & Avoidance of Trauma Cues & 120 & 0.87 & 176 & 1.28 \\
18 & Compensatory Behavior & 159 & 1.16 & 187 & 1.36 \\
19 & Compulsive Behavior & 154 & 1.12 & 258 & 1.88 \\
20 & Diminished Emotional Expression & 320 & 2.33 & 387 & 2.82 \\
21 & Reckless / Self-Destructive Behavior & 151 & 1.10 & 227 & 1.65 \\
22 & Mood / Energy Fluctuation & 345 & 2.51 & 412 & 3.00 \\
23 & Social Anxiety / Avoidance & 342 & 2.49 & 475 & 3.46 \\
24 & Fear of Weight Gain & 182 & 1.33 & 202 & 1.47 \\
25 & Fear of Negative Evaluation & 127 & 0.93 & 157 & 1.14 \\
26 & Flight of Ideas & 179 & 1.30 & 212 & 1.54 \\
27 & Intrusive Symptoms & 138 & 1.01 & 152 & 1.11 \\
28 & Anhedonia / Loss of Interest & 386 & 2.81 & 457 & 3.33 \\
29 & More Talkative & 115 & 0.84 & 182 & 1.33 \\
30 & Obsession & 408 & 2.97 & 564 & 4.11 \\
31 & Panic / Fear & 450 & 3.28 & 528 & 3.85 \\
32 & Pessimism & 302 & 2.20 & 338 & 2.46 \\
33 & Poor Memory & 236 & 1.72 & 275 & 2.00 \\
34 & Sleep Disturbance & 866 & 6.31 & 971 & 7.08 \\
35 & Musculoskeletal Symptoms & 579 & 4.22 & 660 & 4.81 \\
36 & Other Somatic Symptoms & 556 & 4.05 & 644 & 4.69 \\
37 & Sensory Symptoms & 482 & 3.51 & 522 & 3.80 \\
38 & Weight and Appetite Change & 504 & 3.67 & 593 & 4.32 \\
\bottomrule
\end{tabular}
\end{table*}

\begin{table}[t]
\centering
\scriptsize
\setlength{\tabcolsep}{1.5pt}
\renewcommand{\arraystretch}{1.04}
\caption{Per-label annotation statistics for life-event labels.}
\label{tab:lifeevent_label_stats}
\begin{tabular}{@{}rlrrrr@{}}
\toprule
\# & Label & Pos. & Pos.\,[\%] & Mentioned & Mentioned\,[\%] \\
\midrule
1  & Bereavement         & 90    & 0.66  & 196   & 1.43 \\
2  & Health              & 6,030 & 43.94 & 7,409 & 53.99 \\
3  & Relationship Change & 303   & 2.21  & 658   & 4.79 \\
4  & Family Member Change& 46    & 0.34  & 199   & 1.45 \\
5  & Relocation          & 116   & 0.85  & 228   & 1.66 \\
6  & Education           & 236   & 1.72  & 431   & 3.14 \\
7  & Legal               & 35    & 0.26  & 224   & 1.63 \\
8  & Lifestyle Change    & 370   & 2.70  & 562   & 4.10 \\
9  & Social Event        & 106   & 0.77  & 213   & 1.55 \\
10 & Occupation          & 334   & 2.43  & 524   & 3.82 \\
11 & Financial           & 175   & 1.28  & 365   & 2.66 \\
12 & Identity            & 54    & 0.39  & 194   & 1.41 \\
\bottomrule
\end{tabular}
\end{table}

\section{Psychiatric Signal Taxonomy}
\subsection{Symptom Taxonomy}

To support interpretable and cross-lingual psychiatric signal analysis, we constructed a bilingual symptom taxonomy aligned with DSM-5-oriented clinical concepts. The taxonomy covers affective, cognitive, behavioral, somatic, and trauma-related symptom dimensions adapted from prior psychiatric NLP frameworks~\cite{zhang-etal-2022-symptom}. Each Chinese symptom category was manually aligned with its English counterpart to improve semantic consistency across platforms and languages.

Tables~\ref{tab:symptom_taxonomy_1} and~\ref{tab:symptom_taxonomy_2} present the complete symptom taxonomy used in our analyses.

\begin{table}[t]
\centering
\footnotesize
\setlength{\tabcolsep}{2pt}
\renewcommand{\arraystretch}{1.03}
\caption{Complete bilingual symptom taxonomy used in MATCHA (Part 1).}
\label{tab:symptom_taxonomy_1}

\begin{tabularx}{\columnwidth}{@{}>
{\raggedright\arraybackslash}p{0.34\columnwidth}>{\raggedright\arraybackslash}X@{}}
\toprule
\textbf{English Label} &
\textbf{Semantic Gloss} \\
\midrule

Irritable Mood &
Persistent irritability, anger, frustration, low emotional tolerance \\

Anxious Mood &
Excessive worry, panic anticipation, fear of catastrophic outcomes \\

Autonomic Symptoms &
Sweating, accelerated heart rate, dry mouth, shortness of breath \\

Cardiovascular Symptoms  &
Chest pain, palpitations, cardiovascular discomfort \\

Catatonic Behavior &
Motor rigidity, stupor, mutism, abnormal movement behavior \\

Decreased Energy Tiredness Fatigue &
Exhaustion, low energy, persistent tiredness \\

Depressed Mood &
Sadness, hopelessness, emptiness, emotional numbness \\

Gastrointestinal Symptoms &
Nausea, stomach pain, digestive discomfort \\

Genitourinary Symptoms &
Urinary discomfort, reproductive-system-related symptoms \\

Hyperactivity Agitation &
Restlessness, pacing, inability to remain still, agitation \\

Impulsivity &
Impulsive decisions, poor behavioral inhibition \\

Inattention &
Distractibility, forgetfulness, difficulty concentrating \\

Indecisiveness &
Difficulty making decisions, hesitation, uncertainty \\

Respiratory Symptoms &
Shortness of breath, choking sensations, breathing difficulty \\

Suicidal Ideas &
Thoughts of death, suicidal ideation, suicide planning or attempts \\

Worthlessness and Guilty &
Self-blame, guilt, negative self-evaluation, shame \\

Avoidance of Stimuli &
Avoidance of trauma-related memories or external reminders \\

Compensatory Behaviors to Prevent Weight Gain
&
Purging, excessive exercise, compensatory eating behaviors \\

Compulsions &
Checking, cleaning, counting, repetitive compulsive actions \\

\bottomrule
\end{tabularx}
\end{table}

\begin{table}[t]
\centering
\footnotesize
\setlength{\tabcolsep}{2pt}
\renewcommand{\arraystretch}{1.03}
\caption{Complete symptom taxonomy used in MATCHA (Part 2).}
\label{tab:symptom_taxonomy_2}

\begin{tabularx}{\columnwidth}{@{}>{\raggedright\arraybackslash}p{0.34\columnwidth}>{\raggedright\arraybackslash}X@{}}
\toprule
\textbf{English Label} &
\textbf{Semantic Gloss} \\
\midrule

Diminished Emotional Expression &
Reduced facial expression, monotone speech, emotional flattening \\

Do Things Easily Get Painful Consequences &
Risk-taking, self-harm, dangerous impulsive behavior \\

Drastical Shift in Mood and Energy &
Abrupt changes in mood and energy levels \\

Fear About Social Situations &
Fear or avoidance of conversations, strangers, or public performance \\

Fear of Gaining Weight &
Fear of weight gain, body-image disturbance, restrictive eating \\

Fears of Being Negatively Evaluated &
Fear of criticism, embarrassment, or social judgment \\

Flight of Ideas &
Rapid topic shifts, racing thoughts, disorganized thinking \\

Intrusion Symptoms &
Flashbacks, intrusive traumatic memories, trauma-related nightmares \\

Loss of Interest or Motivation &
Loss of motivation, social withdrawal, inability to experience pleasure \\

More Talkative &
Pressured speech, excessive talking, difficult-to-interrupt speech \\

Obsession &
Intrusive thoughts, contamination fears, symmetry obsessions \\

Panic Fear &
Panic attacks, acute fear responses, intense anxiety episodes \\

Pessimism &
Hopelessness, negative expectations about the future \\

Poor Memory &
Forgetfulness, impaired recall, memory difficulties \\

Sleep Disturbance &
Insomnia, hypersomnia, nightmares, reduced sleep need \\

Somatic Muscle &
Muscle pain, body aches, stiffness, movement discomfort \\

Somatic Symptoms Others &
General unexplained physical discomfort or bodily complaints \\

Somatic Symptoms Sensory &
Heat sensations, numbness, tingling, dizziness \\

Weight and Appetite Change &
Binge eating, appetite loss, significant weight change \\

\bottomrule
\end{tabularx}
\end{table}

\subsection{Life-Event Taxonomy}

In addition to symptom-related signals, we annotated major life events frequently associated with mental-health experiences on social media. The life-event taxonomy was adapted from prior work on psychiatric life-event modeling~\cite{lv-etal-2025-tracking,chen-etal-2024-mapping} and manually aligned across Chinese and English settings.

The taxonomy spans multiple dimensions of daily life, including health, relationships, education, work, finances, relocation, and identity-related experiences. Table~\ref{tab:lifeevent_taxonomy} presents representative categories and example sub-events.

\begin{table}[t]
\centering
\footnotesize
\setlength{\tabcolsep}{2pt}
\renewcommand{\arraystretch}{1.03}
\caption{Representative life-event categories in MATCHA.}
\label{tab:lifeevent_taxonomy}
\begin{tabularx}{\columnwidth}{@{}>{\raggedright\arraybackslash}p{0.34\columnwidth}>{\raggedright\arraybackslash}X@{}}
\toprule
\textbf{Category} & \textbf{Representative Sub-events} \\
\midrule

Health &
Hospitalization, medication use, surgery, self-harm, recovery from addiction \\

Relationship Changes &
Breakup, divorce, marriage, family conflict, parenting difficulties \\

Career &
Job loss, promotion, workplace discrimination, unemployment \\

Education &
Examinations, school transition, academic stress, dropping out \\

Financial &
Debt, bankruptcy, major financial difficulty, home purchase \\

Relocation &
Moving cities, immigration, homelessness, leaving parents' home \\

Identity &
Coming out, gender transition, religious or political identity change \\

Death &
Death of spouse, parent, child, friend, or pet \\

Lifestyle Change &
Changes in habits, social activities, recreation, military service \\

New Birth in Family &
Childbirth, adoption, becoming a parent or grandparent \\

Legal &
Arrest, lawsuit, incarceration, loss of driver's license \\

Societal &
Pandemic, war, natural disaster, major political events \\

\bottomrule
\end{tabularx}
\end{table}


\section{Homepage Posts as a Cue-Reduced Testbed}
\label{app:hp_cue_reduced}

\subsection{Experimental Setup}

To evaluate the complementary value of Homepage Posts (HP), we formulate seven binary disorder-versus-control cohort-classification tasks covering ADHD, ASD, BD, DEP, GAD, OCD, and SZ. We compare three input settings: \textbf{STP-only}, using posts retrieved from disorder-specific Super-Topics; \textbf{HP-only}, using users' broader public homepage timelines; and \textbf{STP+HP}, combining both sources.

All experiments use user-level stratified train/validation/test splits and are repeated with five random seeds. We use TF--IDF features with logistic regression and report Macro-F1 and AUROC. To ensure deterministic construction of the STP+HP condition, user IDs are sorted before aggregation and \texttt{PYTHONHASHSEED} is fixed across runs.

\subsection{Cue Prevalence and Cohort Classification}

Table~\ref{tab:hp_cue_prevalence} reports the prevalence of explicit disorder and clinical cues at the post level. Disorder-related cues are substantially more concentrated in STP than HP: 13.88\% of STP contain a disorder-name cue, compared with 3.72\% of HP. Considering either disorder or clinical cues, the prevalence is 19.74\% for STP and 4.64\% for HP.

\begin{table}[t]
\centering
\begingroup
\footnotesize
\setlength{\tabcolsep}{2pt}
\renewcommand{\arraystretch}{1.08}
\caption{Post-level prevalence of explicit disorder and clinical cues across sources.}
\label{tab:hp_cue_prevalence}
\begin{tabular}{@{}lrrrr@{}}
\toprule
Source & Posts & Dis.\,[\%] & Clin.\,[\%] & Either\,[\%] \\
\midrule
HP-control & 153,704   & 0.17  & 0.42 & 0.57 \\
STP        & 169,508   & 13.88 & 8.22 & 19.74 \\
HP         & 1,559,045 & 3.72  & 1.31 & 4.64 \\
\bottomrule
\end{tabular}
\endgroup
\end{table}

Despite the lower cue prevalence, HP remains predictive of cohort membership. Under the original-text condition, HP achieves an average Macro-F1 of 0.842, while STP achieves 0.953. This performance gap is partly expected because STP is collected directly from disorder-specific communities, where explicit topical cues are more concentrated.

\subsection{Cue-Masking Analysis}

We next test whether classification performance is primarily driven by explicit terminology. We apply the same masking procedure symmetrically to STP and HP under three conditions: \textbf{A}, original text; \textbf{B}, disorder-name cues masked; and \textbf{C}, disorder-name and clinical cues masked.

Table~\ref{tab:hp_masking_summary} summarizes the seven-disorder average results. After masking disorder and clinical cues, STP Macro-F1 decreases from 0.953 to 0.936 ($\Delta=-0.017$), whereas HP decreases from 0.842 to 0.833 ($\Delta=-0.009$). Thus, the performance reduction for STP is approximately twice that of HP.


\begin{table*}[t]
\centering
\small
\setlength{\tabcolsep}{5pt}
\renewcommand{\arraystretch}{1.08}
\caption{Cue-masking performance. Panel A compares original and fully
masked text; Panel B reports tiered masking results. Values are mean
$\pm$ standard deviation over five seeds.}
\label{tab:hp_masking_summary}
\begin{tabular}{@{}lrrrrr@{}}
\toprule
\multicolumn{6}{@{}l}{\textit{Panel A: Original versus fully masked text}} \\
Source & F1$_{\mathrm{orig}}$ & F1$_{\mathrm{masked}}$ & $\Delta$F1
& AUROC$_{\mathrm{orig}}$ & AUROC$_{\mathrm{masked}}$ \\
\midrule
STP & $0.953 \pm 0.006$ & $0.936 \pm 0.007$ & $-0.017$ & 0.995 & 0.992 \\
HP  & $0.842 \pm 0.012$ & $0.833 \pm 0.012$ & $-0.009$ & 0.932 & 0.927 \\
\addlinespace[3pt]
\multicolumn{6}{@{}l}{\textit{Panel B: Tiered cue masking}} \\
Condition & A & B & C & $\Delta$(A--B) & $\Delta$(A--C) \\
\midrule
HP-only  & $0.842 \pm 0.012$ & $0.833 \pm 0.012$ & $0.833 \pm 0.012$ & 0.009 & 0.009 \\
STP-only & $0.953 \pm 0.006$ & $0.937 \pm 0.007$ & $0.936 \pm 0.007$ & 0.016 & 0.017 \\
STP+HP   & $0.858 \pm 0.009$ & $0.848 \pm 0.010$ & $0.847 \pm 0.010$ & 0.010 & 0.010 \\
\bottomrule
\end{tabular}
\end{table*}

Table~\ref{tab:hp_masking_per_disorder} reports the per-disorder comparison between the original condition (A) and the most restrictive masking condition (C). The magnitude of the masking-related decrease differs across disorders; the largest STP decreases occur for ADHD (0.043), ASD (0.030), and OCD (0.018).

\begin{table*}[t]
\centering
\small
\setlength{\tabcolsep}{5pt}
\renewcommand{\arraystretch}{1.08}
\caption{Per-disorder Macro-F1 before and after masking disorder-name and clinical cues. $\Delta$ denotes the absolute decrease from the original condition.}
\label{tab:hp_masking_per_disorder}
\begin{tabular}{lcccccc}
\toprule
& \multicolumn{3}{c}{HP} & \multicolumn{3}{c}{STP} \\
\cmidrule(lr){2-4}\cmidrule(lr){5-7}
Disorder & Original & Masked & $\Delta$ & Original & Masked & $\Delta$ \\
\midrule
ADHD & 0.858 & 0.847 & 0.011 & 0.953 & 0.910 & \textbf{0.043} \\
ASD  & 0.874 & 0.856 & 0.018 & 0.967 & 0.937 & \textbf{0.030} \\
BD   & 0.850 & 0.849 & 0.001 & 0.952 & 0.949 & 0.003 \\
DEP  & 0.789 & 0.783 & 0.006 & 0.956 & 0.952 & 0.004 \\
GAD  & 0.808 & 0.805 & 0.002 & 0.977 & 0.972 & 0.005 \\
OCD  & 0.845 & 0.835 & 0.010 & 0.969 & 0.951 & \textbf{0.018} \\
SZ   & 0.868 & 0.856 & 0.012 & 0.897 & 0.883 & 0.014 \\
\midrule
\textbf{Avg.}
& \textbf{0.842} & \textbf{0.833} & \textbf{0.009}
& \textbf{0.953} & \textbf{0.936} & \textbf{0.017} \\
\bottomrule
\end{tabular}
\end{table*}

The tiered masking results in Table~\ref{tab:hp_masking_summary} further show that most of the performance reduction occurs when disorder-name cues are removed. Masking additional clinical terminology produces little further change. This pattern is strongest for STP and substantially weaker for HP.


Together, these results support the complementary role of HP as a cue-reduced evaluation setting. STP contains more concentrated explicit disorder cues and achieves higher absolute classification performance, but is also more sensitive to cue removal. In contrast, HP contains substantially fewer explicit cues while retaining comparable performance after masking. We therefore interpret STP and HP as distinct discourse settings rather than redundant feature sources: STP captures community-centered explicit discourse, whereas HP supports evaluation of broader public-timeline cohort-related signals.

We emphasize that this benchmark predicts membership in MATCHA's community-derived cohorts rather than clinical diagnosis. The results should therefore not be interpreted as evidence that mental disorders can be clinically inferred from ordinary social-media posts.

\section{LLM Annotation Prompt for Identified Positive (IP) Classification}
This section details the prompts used by the LLM. For ADHD and ASD, evidence referring to an affected family member (e.g., a child) is eligible. For BD, DEP, GAD, OCD, and SZ, only evidence referring to the post author is eligible; family-member references do not satisfy the CEP criterion.

\label{sec:prompt}

{\footnotesize

\paragraph{System Instruction.}
\begin{quote}
You are a classifier for disease-related posts.
You should output only a single integer result.
\end{quote}

\paragraph{User Prompt (translated from Chinese).}
\begin{quote}
You will receive a post written by a user in the \{disease\} Super-Topic.

Please determine whether the post contains any of the following information:
\begin{itemize}
    \item The eligible person has been diagnosed with the disease;
    \item The eligible person is taking prescribed medication for the disease;
    \item The eligible person explicitly states that they have the disease;
    \item The eligible person is hospitalized or receiving treatment (e.g., follow-up visits) for the disease.
\end{itemize}

If any of the above conditions are met, output \textbf{1}; otherwise, output \textbf{0}.

Please carefully distinguish whether the post describes a \textbf{real event} that actually happened to the eligible person under the disorder-specific rule above.
If the post does not explicitly mention a confirmed diagnosis, you should \textbf{not} assume the user or their family member has the disease.
\end{quote}

\paragraph{Examples.}
\begin{itemize}
    \item \textit{``I was diagnosed with XX disease last year.''} $\rightarrow$ explicit diagnosis $\rightarrow$ output \textbf{1}.
    \item \textit{``I took my meds again today, hope the next hospital visit goes well.''} $\rightarrow$ medication and treatment $\rightarrow$ output \textbf{1}.
    \item \textit{``The hospital reminds you to pay attention to the following points for patients with XX disease...''} $\rightarrow$ informational content only $\rightarrow$ output \textbf{0}.
    \item \textit{``I've been having headaches and crying a lot recently, what should I do?''} $\rightarrow$ symptoms only, no confirmed diagnosis $\rightarrow$ output \textbf{0}.
    \item \textit{``I really suspect that I have XX disease.''} $\rightarrow$ self-suspected, not confirmed $\rightarrow$ output \textbf{0}.
\end{itemize}

\paragraph{Example of Model Reasoning (translated from Chinese).}
\begin{quote}
The post describes self-perceived changes and coping behaviors but does not include any explicit mention of medical diagnosis, medication, hospitalization, or confirmed self-identification as a patient.
The analysis therefore relies strictly on explicit linguistic cues rather than inference or speculation, and the correct output is \textbf{0}.
\end{quote}

\paragraph{Output Format.}
The model must output \textbf{only one digit} (0 or 1), without any explanation or additional text.
\par}

\section{Additional Cross-lingual Analysis Details}
\label{app:cross_lingual_details}

For the LIWC comparison, all categories are normalized within platform by contrasting disorder users against matched controls. Positive log ratios therefore indicate categories that are relatively more frequent among disorder users, while negative values indicate categories that are relatively less frequent. The main text reports representative categories with the largest and most interpretable deviations.

For the symptom and life-event comparison, the Chinese and English classifiers are trained independently to reduce direct transfer bias across languages. Remaining differences may still reflect annotation granularity, label-inventory mismatch, and platform-specific discourse conventions. Accordingly, all cross-lingual results are interpreted as aggregate distributional trends rather than clinical or individual-level claims.

\end{document}